\documentclass[letterpaper, 10 pt, conference]{ieeeconf}  

\IEEEoverridecommandlockouts                              

\usepackage{amsmath}    
\usepackage{graphicx}   
\usepackage[lofdepth,lotdepth]{subfig}
\usepackage{tabularx}
\usepackage[font=footnotesize,labelfont=footnotesize]{caption}
\usepackage{verbatim}   
\usepackage{color}      
\usepackage{hyperref} 
\usepackage{multirow}
\usepackage{multicol}
\usepackage{graphicx}
\usepackage{rotating}
\usepackage{array}
\usepackage{xspace}
\usepackage{indentfirst}
\usepackage{amssymb}
\usepackage{float}
\usepackage{algorithm}
\usepackage[algo2e]{algorithm2e}
\usepackage{algorithmic}
\usepackage{comment}
\usepackage{sidecap}
\usepackage{booktabs}
\usepackage{breqn}
\usepackage{cite}
\usepackage{mathtools}
\usepackage{graphicx}
\graphicspath{{./figures/}}
\usepackage{caption}
\usepackage{float}
\usepackage{lipsum}
\usepackage{xcolor}
\usepackage{siunitx}
\usepackage{xspace}
\usepackage{bm}
\usepackage{gensymb}

\usepackage{paralist}

\DeclareMathOperator*{\argmax}{argmax} 

\newcommand{\gobble}[1]{}

\newcounter{tecounter}
\title{\LARGE \bf
Localization of a Smart Infrastructure Fisheye Camera in a Prior Map for Autonomous Vehicles
}

\author{Subodh Mishra$^{*}$, Armin Parchami, Enrique Corona, Punarjay Chakravarty, \\Ankit Vora, Devarth Parikh, and Gaurav Pandey
\thanks{$^{*}$Graduate student in the Department of Mechanical Engineering, Texas A\&M University
        {\tt\small subodh514@tamu.edu}, work done as intern}%
\thanks{All other authors with the Ford Autonomous Vehicles LLC, USA}%
}

\begin{document}
\maketitle
\thispagestyle{empty}
\pagestyle{empty}
\begin{abstract}
 This work presents a technique for localization of a smart infrastructure node, consisting of a fisheye camera, in a prior map. These cameras can detect objects that are outside the line of sight of the autonomous vehicles (AV) and send that information to AVs using V2X technology. However, in order for this information to be of any use to the AV, the detected objects should be provided in the reference frame of the prior map that the AV uses for its own navigation. Therefore, it is important to know the accurate pose of the infrastructure camera with respect to the prior map. Here we propose to solve this localization problem in two steps, \textit{(i)} we perform feature matching between perspective projection of fisheye image and bird's eye view (BEV) satellite imagery from the prior map to estimate an initial camera pose, \textit{(ii)} we refine the initialization by maximizing the Mutual Information (MI) between intensity of pixel values of fisheye image and reflectivity of 3D LiDAR points in the map data. We validate our method on simulated data and also present results with real world data.
\end{abstract}

\begin{keywords} Fisheye Camera, Camera Localization, Mutual Information
\end{keywords}

\section{Introduction} 
\label{sec: introduction}
\begin{figure}[!ht]
\centering
\includegraphics[width=0.4\textwidth]{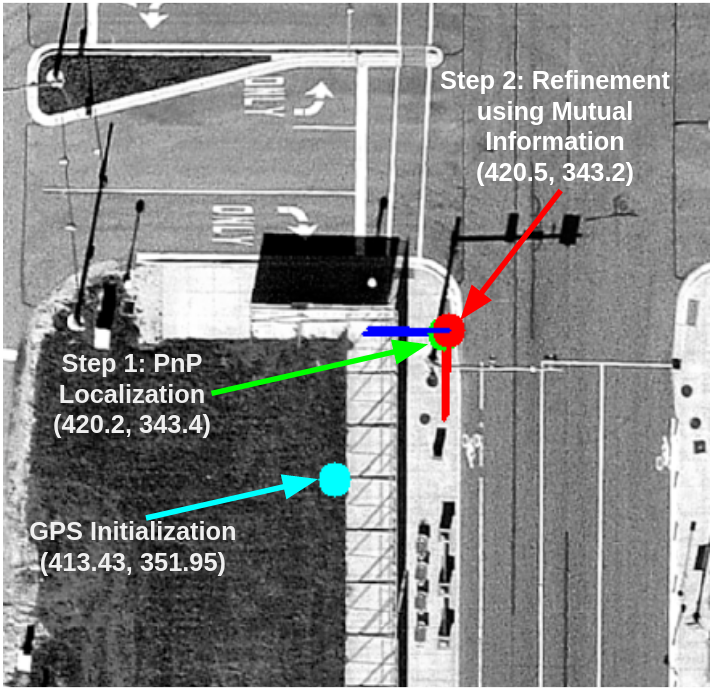}
\caption{Overview of our two step approach to camera localization. The coordinates shown are in meter, wrt the origin of the map. We start with a noisy GPS initialization (cyan), then perform feature matching between satellite image and rectified fisheye image to obtain an inital camera pose - PnP Localization (Green), and finally maximize the Mutual Information between LiDAR map and the fisheye image to obtain a refined camera localization estimate.}
\label{fig: cameraregistrationinpriormap}
\vspace{-20pt}
\end{figure}

Environment perception in Autonomous Vehicles (AV) is a challenging problem. With the current approach of using only on-board sensors to solve the perception problem, it is impossible to sense occluded areas and mitigate the effects of sensor outage. Complex traffic intersections with buildings close to the curb may minimize the field of view of an AV's sensors. Integration of smart infrastructure nodes (sensing and compute) on roads where AVs operate can help overcome these challenges. The elevated and static view-point of the smart sensors enables them to observe the environment, detect more objects in the scene, and communicate that information to AVs. AVs can fuse that information with their own sensor measurements and augment their situational awareness. Fisheye cameras are well known for their low cost and wide Field of View (FoV), making them suitable for such smart infrastructure based sensing applications. However, the fisheye camera needs to provide this information in the same coordinate frame as the vehicle. For this reason, they need to be localized or registered within the same map that is used by the AV for navigation. In this work, we propose a method to localize a downward looking static smart infrastructure fisheye camera in a prior map consisting of a metric satellite image, and a co-registered LiDAR map of ground points with their LiDAR reflectivity values. An overview of the approach is shown in Figure \ref{fig: cameraregistrationinpriormap}.

\section{Related Work}
\label{sec: relatedwork}
There are several contributions on localization of camera images in prior maps (satellite imagery or LiDAR generated 3D maps). Satellite maps can be procured easily from third party sources \cite{vora2020aerial, maxar}, and with the widespread use of LiDARs in autonomous driving, we now have several high definition map providers which provide dense 3D map of the environment \cite{vora2019high}. This has made localization of cheaper sensors like cameras on prior maps an active area of research with an ultimate aim of using cheaper sensors on-board an AV.  \cite{Yu2020MonocularCL} presents monocular perspective camera localization in pre-built 3D LiDAR maps using 2D-3D line correspondences. This method shows promise only in structured environment where lines can be easily detected in both camera and the LiDAR map. \cite{Wolcott2014VisualLW} presents LiDAR map based monocular camera localization in urban environment. Unlike \cite{Yu2020MonocularCL} which depends on detection of geometric primitives like lines in both the sensing modalities, \cite{Wolcott2014VisualLW} uses dense appearance based approach which work in unstructured environments.  In \cite{Wolcott2014VisualLW}, given an initial belief of the camera pose, they generate several synthetic views of the environment by projecting the LiDAR map points using a perspective camera model, and compare these synthetic views against the live camera feed. The synthetic view which maximizes the Normalized Mutual Information (NMI) between the real image gray scale values and the projected points' LiDAR reflectivity values, is the solution to the localization problem. \cite{Wolcott2014VisualLW} draws inspiration from work done on 3D-LiDAR Camera extrinsic calibration described in \cite{MIGP}, which uses maximization of Mutual Information (MI) for calibrating a 3D-LiDAR Camera pair. \cite{Viswanathan} presents a camera localization technique which matches ground imagery obtained by cameras onboard an AV to the available satellite imagery. The camera images are warped to obtain a bird's eye view (BEV) of the ground. Next, the BEV image is matched with the given satellite imagery using using SIFT \cite{Lowe:2004:DIF:993451.996342} features. 

While the above mentioned approaches provide solutions for perspective cameras, our focus is to localize (estimate $[\mathbf{^{C}R_W}, \mathbf{^{C}t_W}]$ in Equation \ref{eqn: fisheyeprojectionmodel}) a downward looking static fisheye camera (Figure \ref{fig: fisheyeimage}) in a prior map (Figure \ref{fig: priormap}) which consists of 2D satellite imagery with metric information (Figure \ref{fig: satimage}) and 3D LiDAR map of ground points (Figures \ref{fig: intensityimage}, \ref{fig: zheightimage}). We initialize the camera localization using feature matching as done in \cite{Viswanathan}, and refine the camera localization using maximization of a MI based cost function as done in \cite{Wolcott2014VisualLW} and \cite{MIGP}.

\section{Overview}
\label{sec: overview}
In this section we provide an overview of the various components of our implementation.
\subsection{Fisheye Camera}
\label{sec: fisheyecamera}
\vspace{-30pt}
\begin{figure}[H]
  \centering
  \subfloat[Fisheye image captured at a traffic intersection in our simulator]{\includegraphics[width=0.225\textwidth]{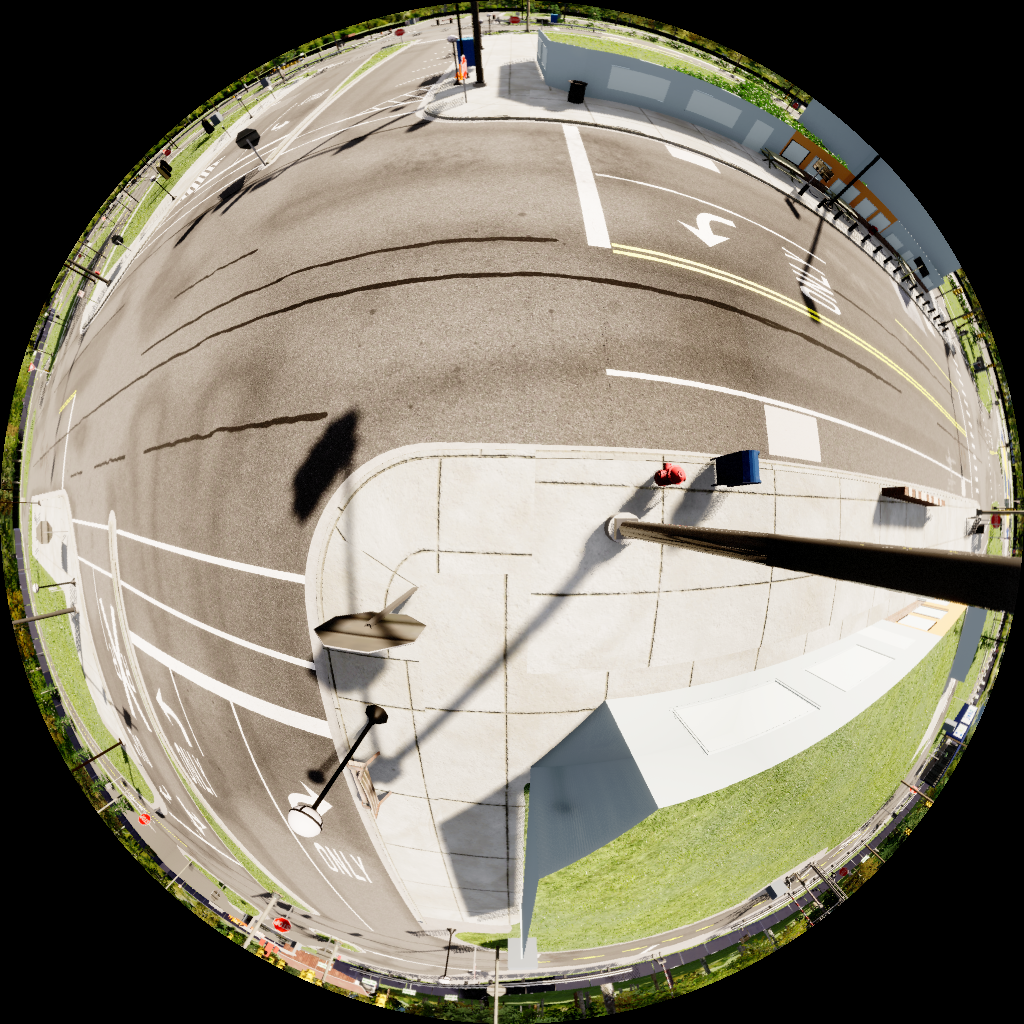}\label{fig: fisheyeimage}}
  \quad
  \subfloat[Rectification of fisheye image to perspective image]{\includegraphics[width=0.225\textwidth]{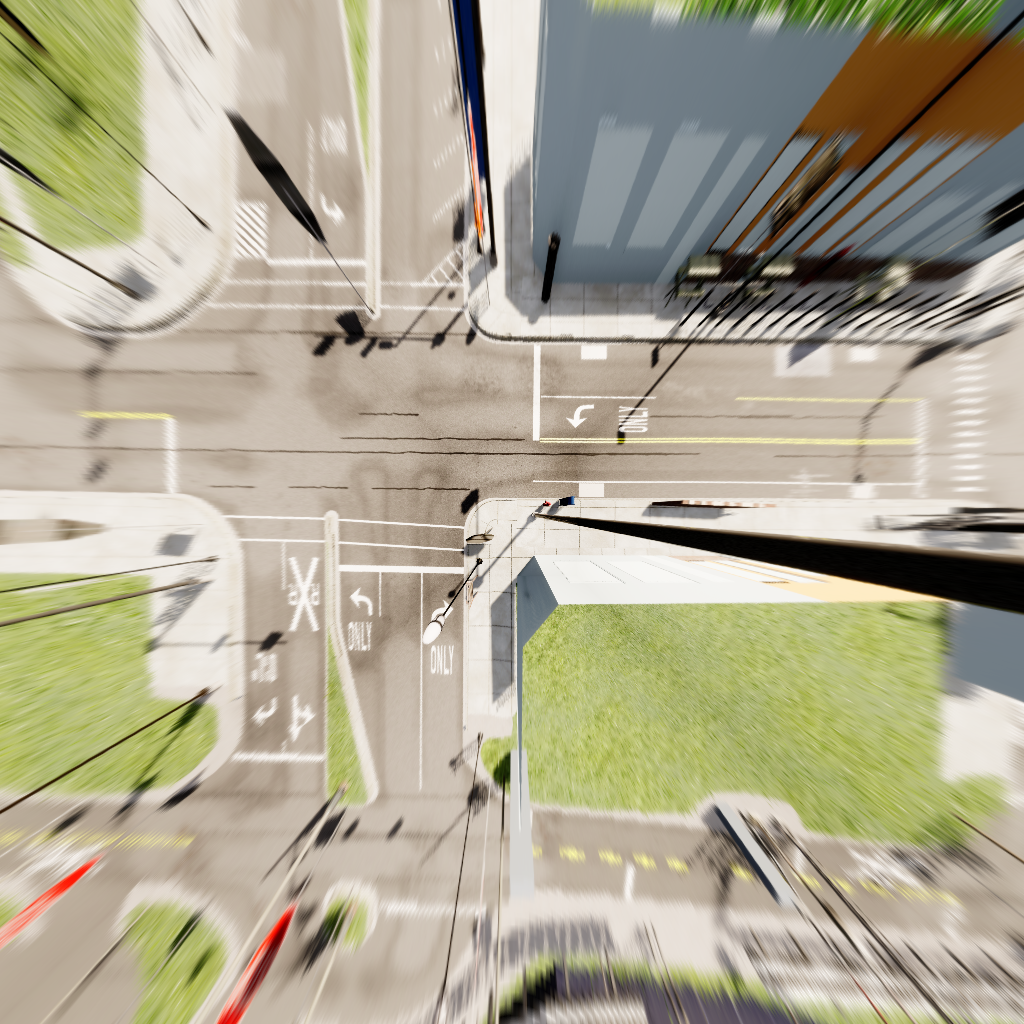}\label{fig: perspectifiedfisheyeimage}}
  \caption{Fisheye Image (Figure \ref{fig: fisheyeimage}) and its perspective rectification (Figure \ref{fig: perspectifiedfisheyeimage})} 
  \label{fig: fisheyevsperspectiveprojection}
\vspace{-10pt}
\end{figure}
We use the fisheye camera projection model proposed in \cite{MeiRives} to project 3D point $\mathbf{P_{W}} = [X_W, Y_W, Z_W]$ defined in the prior map coordinate frame $\mathbf{W}$ to a 2D point $\mathbf{p}$ on the fisheye camera image plane using Equation \ref{eqn: fisheyeprojectionmodel}.
\begin{align}
    \textbf{p} &= \mathbf{\Pi}(K, D, \xi, [\mathbf{^{C}R_W}, \mathbf{^{C}t_W}], \mathbf{P_{W}})
    \label{eqn: fisheyeprojectionmodel}
\end{align}
Here $\mathbf{\Pi}()$ is the projection function, $K=\begin{bmatrix}
f_x & s & c_x \\
0 & f_y & c_y \\
\end{bmatrix}$, $D = [k_1, k_2, p_1, p_2]$ and $\xi$ are the camera intrinsics, and $[\mathbf{^{C}R_W}$, $\mathbf{^{C}t_W}]$ is the camera extrinsic. $\mathbf{^{C}R_W}$ $\in SO(3)$ is an orthonormal rotation matrix and  $\mathbf{^{C}t_W}$ $\in R^{3}$ is a 3D vector. The goal of this work is to estimate the unknown camera extrinsic $[\mathbf{^{C}R_W}$, $\mathbf{^{C}t_W}]$ in the prior map frame $\mathbf{W}$. 
\subsubsection{Intrinsic Calibration} \label{sec: intrinsiccalibration} We estimate the intrinsic parameters $K$, $D$ and $\xi$ by collecting several images of a large checkerboard at different poses, and feeding those images to the omnidirectional camera calibrator in OpenCV\cite{opencv_library}, which provides an implementation\footnote{\url{https://docs.opencv.org/4.5.2/dd/d12/tutorial_omnidir_calib_main.html}} of the intrinsic calibration technique presented in \cite{boli}.

\subsubsection{Rectification} \label{sec: perspectiverectification} The intrinsic camera calibration parameters are used to rectify the fisheye image (Figure \ref{fig: fisheyeimage}) into its corresponding perspective image (Figure \ref{fig: perspectifiedfisheyeimage}) utilizing OpenCV's rectification routines. Although perspective rectification results in loss of field of view, it makes the application of computer vision algorithms developed for perspective images possible for fisheye images.

\subsection{Prior Map}
\label{sec: priormap}
\begin{figure*}[!ht]
  \centering
  \subfloat[Satellite Map]{\includegraphics[width=0.3\textwidth]{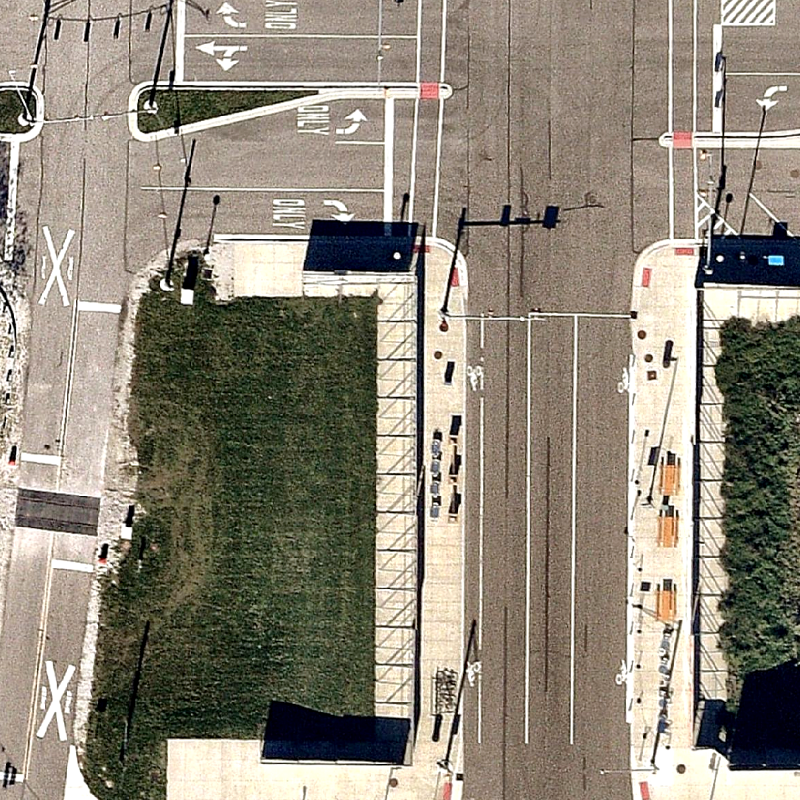}\label{fig: satimage}}
  \quad
  \subfloat[LiDAR Map: Reflectivity of Ground Points]{\includegraphics[width=0.3\textwidth]{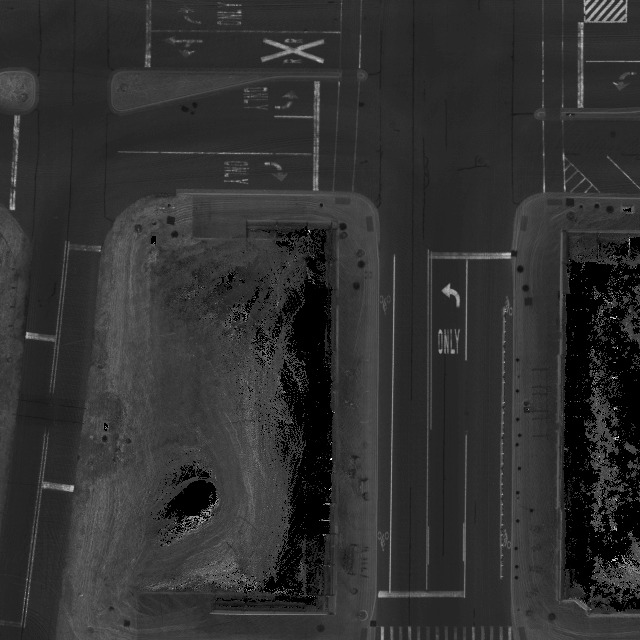}\label{fig: intensityimage}}
  \quad
  \subfloat[LiDAR Map: Height of Ground Points]{\includegraphics[width=0.3\textwidth]{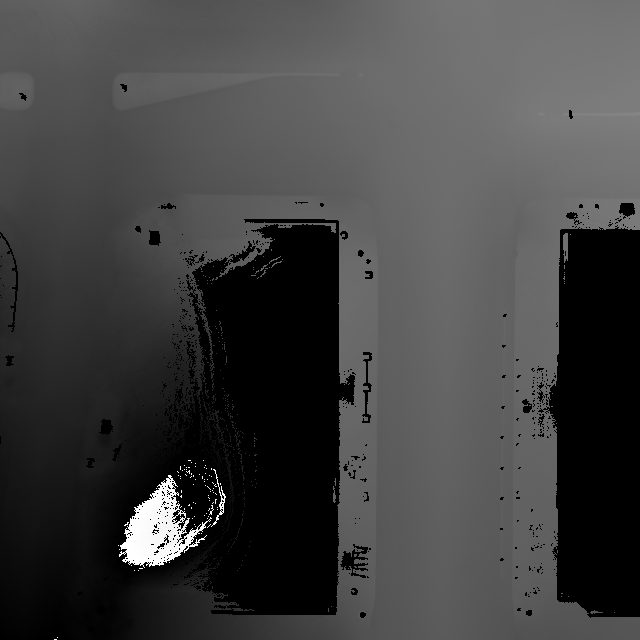}\label{fig: zheightimage}}
  \caption{\textbf{Prior Map:} Figures \ref{fig: satimage}, \ref{fig: intensityimage} $\&$ \ref{fig: zheightimage} show the components of our prior map. The full map is not shown in the interest of space.} 
  \label{fig: priormap}
\end{figure*}

The prior map (Figure \ref{fig: priormap}) consists of two important components which are registered and expressed in the frame of reference $\mathbf{W}$. They are:
\subsubsection{Satellite Map} \label{sec: satmap}
The satellite map is a metric Bird's Eye View (BEV) satellite image (Figure \ref{fig: satimage}). In our case, a pixel on the image corresponds to 0.1 m on the ground. 

\subsubsection{LiDAR Map} \label{sec: LiDARmap}
The prior LiDAR map is built using an offline mapping process described in \cite{Wolcott2014VisualLW}. Broadly, a survey vehicle equipped with several 3D LiDAR scanners and a high end inertial navigation system is manually driven and sensor data is collected in the environment we want to map. Next, an offline pose-graph optimization SLAM (Simultaneous Localization and Mapping) problem is solved to obtain the accurate global pose of the vehicle. Finally, a dense ground point mesh is constructed from the optimized pose graph using region growing techniques which gives a dense 3D point cloud map. The ground points from this dense 3D cloud are used to generate the LiDAR ground reflectivity image (Figure \ref{fig: intensityimage}) and ground height image (Figure \ref{fig: zheightimage}). The LiDAR Map (Figures \ref{fig: intensityimage} and \ref{fig: zheightimage}) is aligned with the satellite imagery (Figure \ref{fig: satimage}) using the GPS measurements from the inertial navigation system.

\section{Problem Formulation}
\label{sec: problemformulation}
\begin{figure*}[!ht]
\centering
\includegraphics[width=\textwidth]{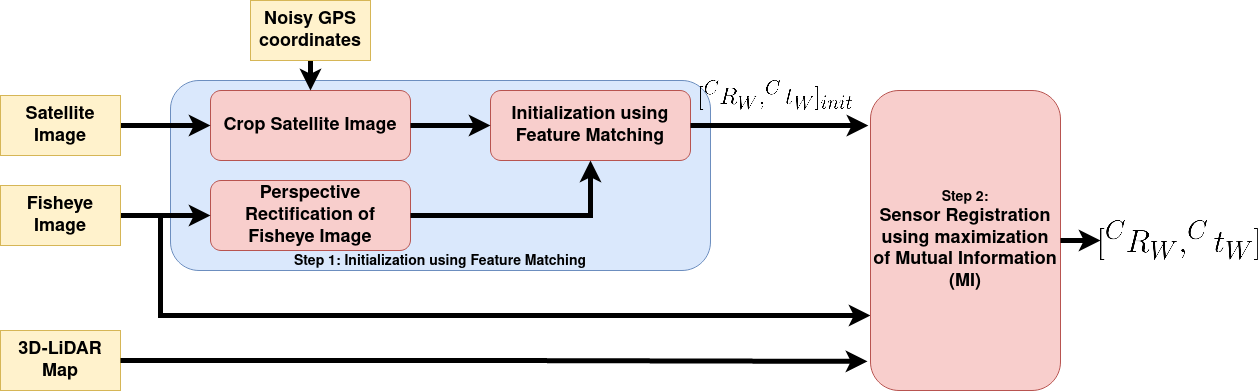}
\caption{\textbf{Overview of the method:} The block diagram shows the two steps involved in our approach. In Step 1, we match features between the perspective projection of fisheye image and a cropped satellite map to initialize camera pose, and in Step 2 we refine this initialization by maximization of Mutual Information between fisheye image and prior 3D-LiDAR map.}
\label{fig: fisheyeregistrationblockdiagram}
\vspace{-20pt}
\end{figure*}
The goal of this work is to estimate the unknown camera pose $[\mathbf{^{C}R_W}$, $\mathbf{^{C}t_W}]$ in the prior map frame $\mathbf{W}$. We assume that we have a noisy estimate of the fisheye camera's (GPS) position (no orientation) in the map which helps us reduce the search space in the prior map. We follow a two step approach to register the camera in the prior map, the details of which are presented in Sections \ref{sec: initsparsefeaturematching} and \ref{sec: refinement_of_localization}, and a broad overview is provided in Figure \ref{fig: fisheyeregistrationblockdiagram}.

\subsection{Initialization using sparse feature matching} 
\label{sec: initsparsefeaturematching}
\begin{figure}[!ht]
  \centering
  \includegraphics[width=0.45\textwidth]{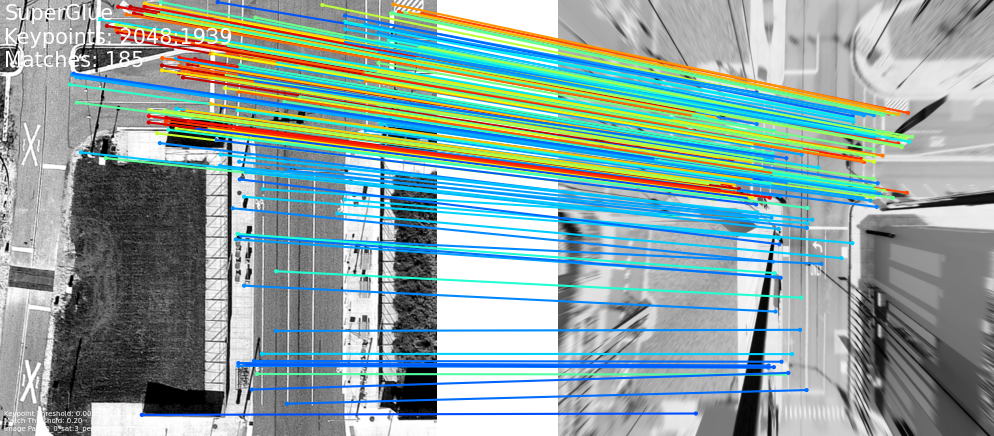}
  \caption{SuperGlue \cite{sarlin20superglue} Matching between Satellite Image (Left) $\&$ perspective projection of fisheye image (Right)} 
  \label{fig: featurematches}
  \vspace{-15pt}
\end{figure}

Traditionally available feature detection, description and matching techniques are usually suitable for perspective images only. Therefore, we rectify the fisheye image into the corresponding perspective image as explained in Section \ref{sec: perspectiverectification}, and use SuperGlue \cite{sarlin20superglue}, a pre-trained deep learning based feature matching algorithm, for matching features (Figure \ref{fig: featurematches}) between the rectified fisheye image and the cropped satellite image (cropped using GPS initialization, refer Figure \ref{fig: fisheyeregistrationblockdiagram}). The matched features are used to solve a Perspective-n-Point (PnP) problem \cite{P3P, RANSAC} to estimate the initial pose of camera (also called the PnP estimate) in the prior map reference frame $\mathbf{W}$. As we know the metric scale of the satellite image (1 pixel  = 0.1 m), we obtain the camera pose in metric units. 
\subsection{Refinement of camera localization using Maximization of Mutual Information}
\label{sec: refinement_of_localization}

Mutual Information (MI) has been used in several fields for registering data from multi-modal sensors \cite{paulviola, MaesMI}. We refine the initial camera pose estimate from Section \ref{sec: initsparsefeaturematching} by maximizing the Mutual Information (MI) between the LiDAR reflectivity of ground points and the fisheye grayscale values at the pixel locations onto which the LiDAR points are projected using the camera pose $[\mathbf{^{C}R_W}, \mathbf{^{C}t_W}]$.

\subsubsection{Theory} MI (Equation \ref{eqn: mutualinformationdefinition}) provides a way to statistically measure mutual dependence between two random variables $X$ and $Y$.
\begin{equation}
    MI(X, Y) = H(X) + H(Y) - H(X, Y) 
    \label{eqn: mutualinformationdefinition}
\end{equation}

Where $H(X)$ and $H(Y)$ are the Shannon entropy over random variables $X$ and $Y$ respectively, and $H(X, Y)$ is the joint Shannon entropy over the two random variables:
\begin{align}
    H(X) &= - \sum_{x \in X} p_{X}(x) \log p_{X}(x) \label{eqn: marginalX}\\
    H(Y) &= - \sum_{y \in X} p_{Y}(y) \log p_{Y}(y) \label{eqn: marginalY}\\
    H(X, Y) &= - \sum_{x \in X} \sum_{y \in Y} p_{X, Y}(x, y) \log p_{XY}(x, y) \label{eqn: jointXY}
\end{align}

The entropy $H(X)$ of a random variable $X$ denotes the amount of uncertainty in $X$, whereas $H(X, Y)$ is the amount of uncertainty when the random variables $X$ and $Y$ are co-observed. The formulation of MI in Equation \ref{eqn: mutualinformationdefinition} shows that maximization of $MI(X, Y)$ is achieved by minimization of the joint entropy $H(X, Y)$, which coincides with minimization of dispersion of two random variable's joint histogram.

\subsubsection{Mathematical Formulation} 
\label{sec: mathematicalformulation}



Let $\{\mathbf{P_{W_{i}}}; i = 1,2, \dots, n\}$ be the set of 3D points whose coordinates are known in the prior map reference frame $\mathbf{W}$ and let $\{X_{i}; i = 1,2, \dots, n\}$ be the corresponding reflectivity values for these points ($X_i \in [0, 255]$). Equation \ref{eqn: fisheyeprojectionmodel} presents the relationship between $\mathbf{P_{W_{i}}}$ and its image projection $\mathbf{p_{i}}$ as a function of $[\mathbf{^{C}R_W}, \mathbf{^{C}t_W}]$.
Let $\{Y_i; i = 1, 2, \dots, n\}$ be the grayscale intensity of the pixels $\mathbf{p_{i}}$ where ${\mathbf{P_{W_{i}}}}$ project onto, such that: 
\begin{align}
    Y_i = I(\mathbf{p_{i}})
\end{align}
where $Y_i \in [0, 255]$ and $I$ is the grayscale fisheye image. Therefore, $X_i$ is an observation of the random variable $X$, and for a given $[\mathbf{^{C}R_W}, \mathbf{^{C}t_W}]$, $Y_i$ is an observation of random variable $Y$. The marginal ($p_{X}(x)$, $p_{Y}(y)$) and joint ($p_{X, Y}(x, y)$) probabilities  of the random variables $X$ and $Y$, required for calculating MI (Equation \ref{eqn: mutualinformationdefinition}), can be estimated using a normalized histogram (Equation \ref{eqn: pdfdefinition}):
\begin{align}
\hat{p}(X = k) = \frac{x_k}{n}, k \in [0, 255]
\label{eqn: pdfdefinition}
\end{align}
where $x_k$ is the observed counts of the intensity value $k$.



\subsubsection{Global Optimization}
$\mathbf{^CR_W} \in SO(3)$ is an orthonormal rotation matrix which can be parameterized as Euler angles $[\phi, \theta, \psi]^{\top}$ and $\mathbf{^{C}t_{W}} = [x, y, z]^{\top}$ is an Euclidean 3-vector. $\psi$ is the rotation of the camera along its principal axis. In our context, the fisheye camera is facing vertically downward so we do not refine the $\phi$ $\&$ $\theta$ and leave it at what the feature matching based technique (Section \ref{sec: initsparsefeaturematching}) determines it to be, which is very close to 0. Therefore, as far as rotation variables are concerned, we refine only $\psi$. We represent all the variables to be optimized together as $\Theta = [x, y, z, \psi]^{\top}$. The optimization is posed as a maximization problem:
\begin{align}
    \hat{\Theta} = \argmax_{\Theta} MI(X, Y; \Theta)
    \label{eqn: globaloptimization}
\end{align}


\section{Experiments and Results}
\label{sec: experimentsandresults}
This section describes the experiments performed to evaluate the proposed technique using data obtained from both our simulator and real world sensor.

\subsection{Simulation Studies}
\label{sec: simulationstudies}
We first validate our approach on a simulator which is built using data from real sensors. The Mathworks' tool RoadRunner \cite{roadrunner} is used to generate the 2D features like lane geometry and lane markings with the satellite map used as a reference. The 3D structures are created using the Unreal Engine Editor \cite{unrealeditor} with the help of real satellite and 3D LiDAR maps. Since the simulated environment is created using the prior map components, it can safely be assumed that the simulator aligns with the real world to a high degree of accuracy. In order to generate the fisheye images, we model a fisheye camera in Unreal Engine using the equidistant model with a field of view of 180 degrees. We demonstrate our approach in simulation for the fisheye image shown in Figure \ref{fig: fisheyeimage}. The fisheye image is first rectified (Section \ref{sec: perspectiverectification}) to generate Figure \ref{fig: perspectifiedfisheyeimage}, which is used for estimating the initial camera pose using the approach in Section \ref{sec: initsparsefeaturematching}. Next, the initialization is refined using maximization of MI (Section \ref{sec: refinement_of_localization}). 
\vspace{-20pt}
\begin{figure}[!ht]
  \centering
  \subfloat[]{\includegraphics[width=0.225\textwidth]{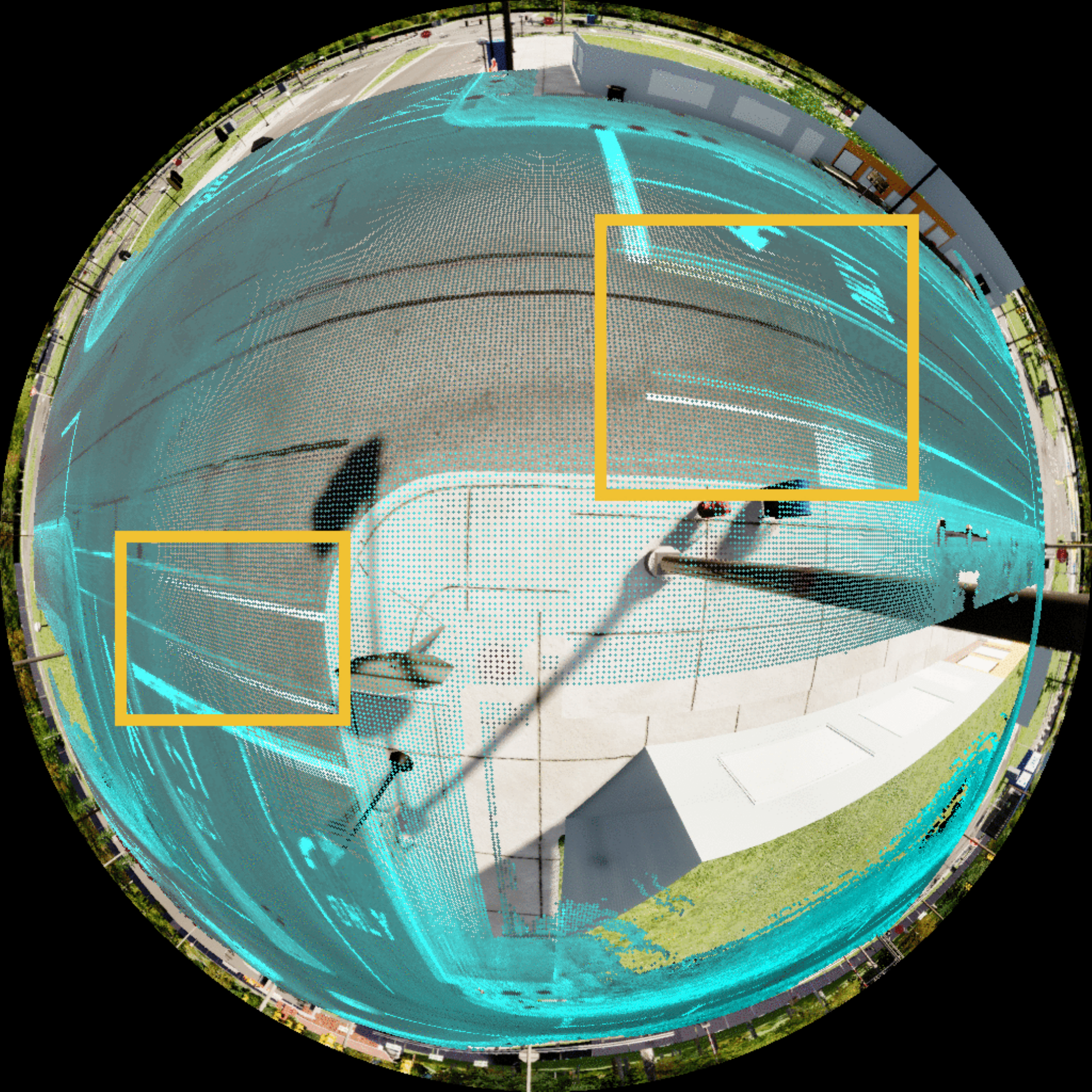}\label{fig: loc1_ht1_pnp_marked}}
  \quad
  \subfloat[]{\includegraphics[width=0.225\textwidth]{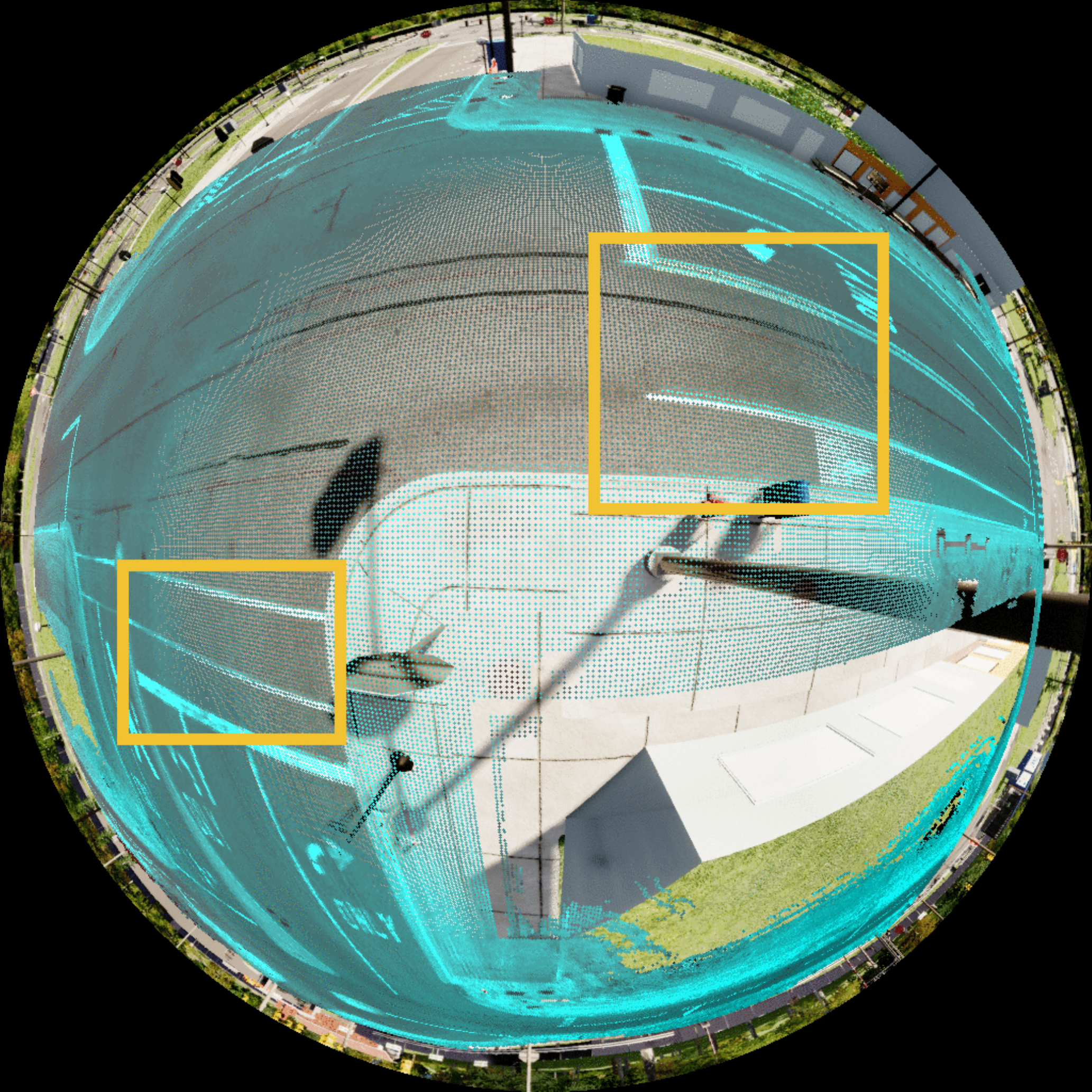}\label{fig: loc1_ht1_mi}}
  \caption{Two step approach for camera localization in prior map : Figure \ref{fig: loc1_ht1_pnp_marked} shows the projection of 3D-LiDAR ground points (cyan) using the initial estimate obtained using feature matching (PnP Estimate from Section \ref{sec: initsparsefeaturematching}), and Figure \ref{fig: loc1_ht1_mi} shows the projection of 3D-LiDAR ground points using the refined camera pose estimate from maximization of MI (from Section \ref{sec: refinement_of_localization}). The misalignment visible in Figure \ref{fig: loc1_ht1_pnp_marked}, is absent in Figure \ref{fig: loc1_ht1_mi} (best viewed digitally).}
  \label{fig: improvementwithMIbasedlocalization}
  \vspace{-15pt}
\end{figure}

\begin{figure}[!ht]
\centering
\includegraphics[width = 0.4\textwidth]{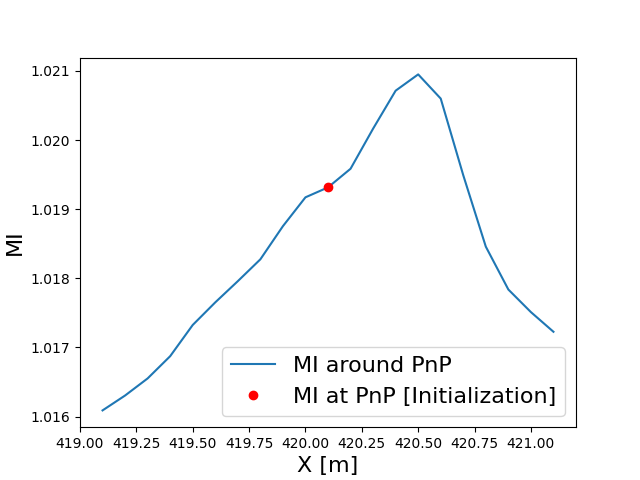}
\caption{
Plot of MI around the PnP estimate (from Section \ref{sec: initsparsefeaturematching}). Plot shows that MI is not at maximum at the PnP estimate, therefore the maximization of MI may reduce the misalignment in projection of 3D-LiDAR points visible in Figure \ref{fig: loc1_ht1_pnp_marked}}\label{fig: xperturbatioMI} 
\end{figure}

\begin{figure}[!ht]
\centering
\includegraphics[width=0.4\textwidth]{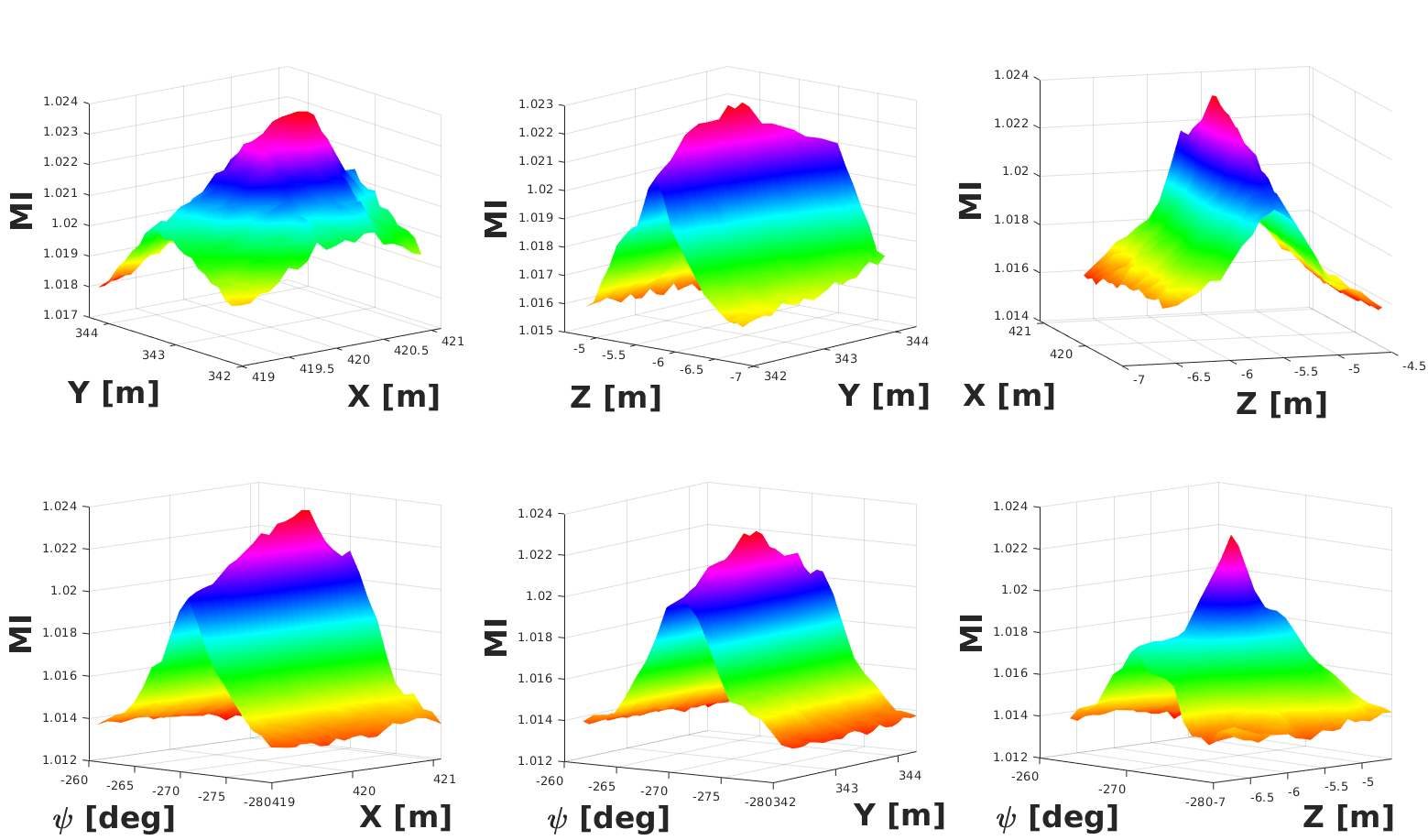}
\caption{\textbf{Plot of MI by perturbing two degrees of freedom around the PnP Estimate} (i.e. the initial estimate from Section \ref{sec: initsparsefeaturematching}). The cost function from single Image LiDAR Scan pair is not differentiable at several points. Hence, we use an exhaustive grid search around the initialization point to arrive at solution where MI is maximized.}
\label{fig: costfunction2d}
\vspace{-10pt}
\end{figure}

We qualitatively validate the camera localization (Figure \ref{fig: improvementwithMIbasedlocalization}) estimate ($[\mathbf{^{C}R_W}, \mathbf{^{C}t_W}]$) by projecting points from the 3D-LiDAR map (Figures \ref{fig: intensityimage} and \ref{fig: zheightimage}) onto the fisheye image (Figure \ref{fig: fisheyeimage}). As shown in Figure \ref{fig: loc1_ht1_pnp_marked}, the projection of LiDAR map points on the fisheye image obtained using the initial camera pose are not well aligned. 
When we plot (Figure \ref{fig: xperturbatioMI}) the MI  around the initial camera pose, we observe that it is not at its maximum at the initial estimate (also called PnP Estimate), thus holding the promise for further improvement. Similarly, Figure \ref{fig: costfunction2d} presents the surface plot of MI, which shows the presence of a global maximum in each sub-plot. Therefore, on solving the optimization problem posed in Equation \ref{eqn: globaloptimization} we obtain camera pose estimate which maximizes the MI between the two modalities and results in negligible misalignment of the projected LiDAR map points in Figure \ref{fig: loc1_ht1_mi}.

\begin{figure}[!ht]
\centering
\includegraphics[width=0.4\textwidth]{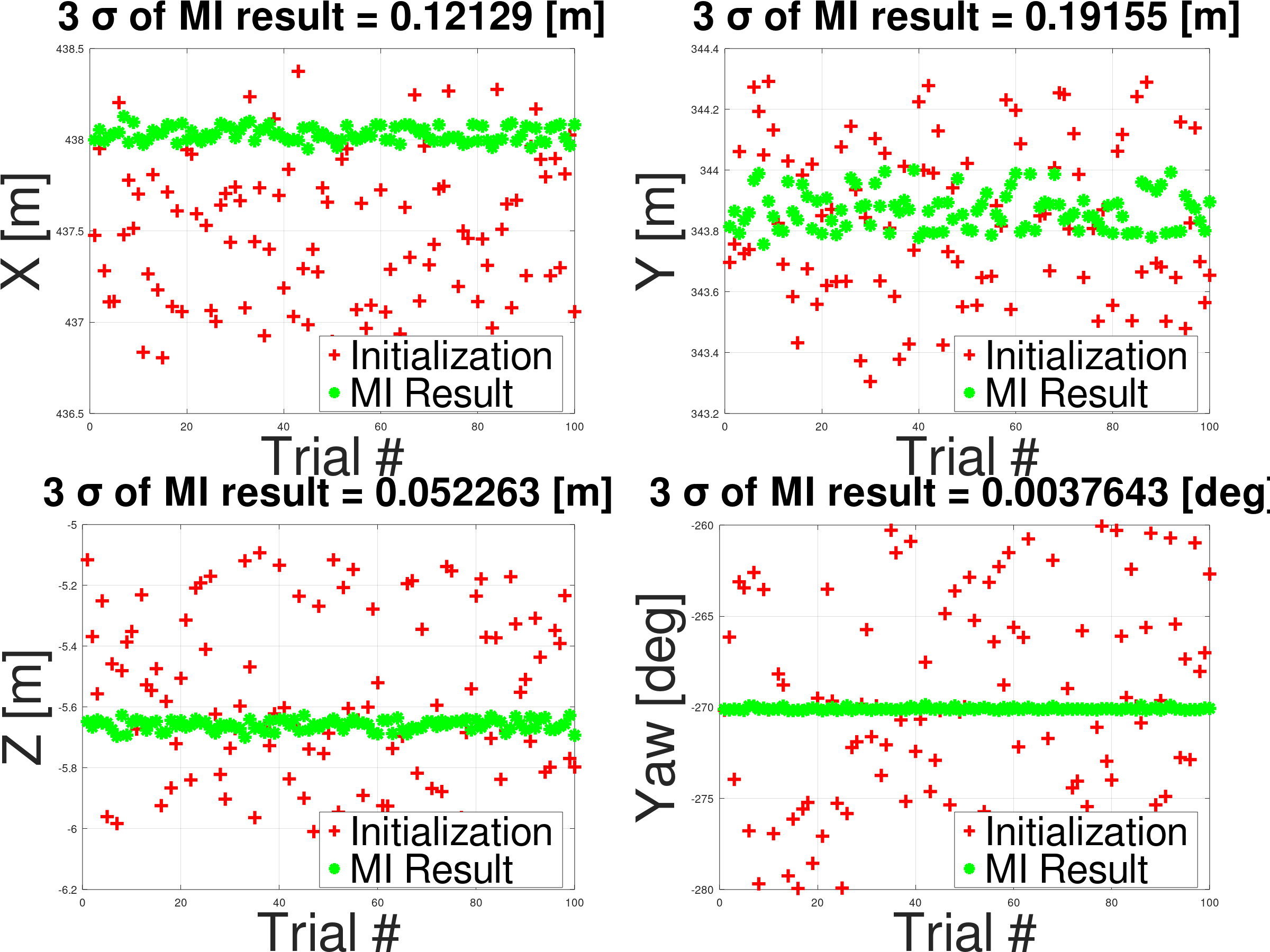}
\caption{Performance of MI based fisheye camera localization refinement for different initial conditions. Here we perform 100 independent trials. \textcolor{red}{+} is the initialization, \textcolor{green}{*} is the final result.}
\label{fig: mi_randominitializaton}
\vspace{-10pt}
\end{figure}

We run 100 independent trials to evaluate the robustness of the MI based refinement method (Section \ref{sec: refinement_of_localization}) to change in initialization (Figure \ref{fig: mi_randominitializaton}). The translation parameters show greater variance when compared to yaw $\psi$. The higher variance of the translation variables can be attributed to the fact that the MI based cost function is less sensitive to changes in translation variables, especially in the outdoor scenario where most of the points lie in the far field.
In the limiting case, far away points are considered to be points at infinity, represented as $[x, y, z, 0]^{T}$, which, under camera projection (Equation \ref{eqn: fisheyeprojectionmodel}), render the translation variable ($\mathbf{^{C}t_{W}}$) in the optimization problem (Equation \ref{eqn: globaloptimization}) un-observable. 
This result is also presented in \cite{MIGP}, specifically when discussing sensor registration in an outdoor environment using only a single image - LiDAR scan pair, which is similar to our situation. 

\subsection{Real World Experiments}
In order to demonstrate the validity of our algorithm in realistic situations, we conducted experiments with data collected from a real fisheye camera (Figure \ref{fig: realfisheyedatacollection}).
\subsubsection{System Description} 
\begin{figure}[H]
\centering
\includegraphics[width=0.4\textwidth]{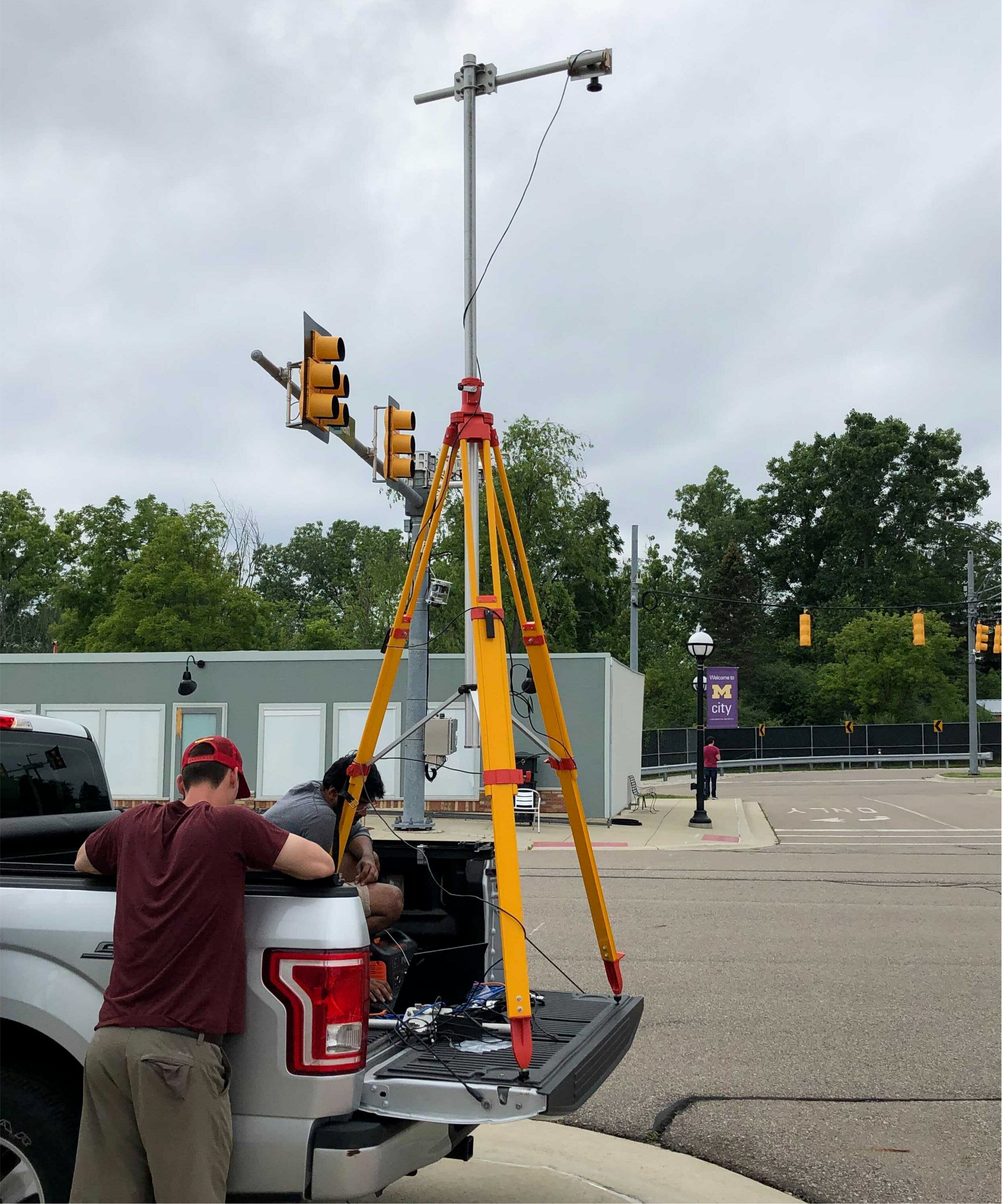}
\caption{Collecting data for real experiments using a tripod mounted downward looking fisheye camera. Our ultimate goal is to mount these cameras, along with our smart infrastructure nodes, at challenging intersections for navigation of autonomous vehicles.}
\label{fig: realfisheyedatacollection}
\vspace{-15pt}
\end{figure}

We use a fisheye lens Fujinon FE185C057HA 2/3 inch sensor, which provides $185^{\circ}$ of vertical and horizontal FoV. Our camera is a 5MP Sony IMX264. We mount our sensor from a tripod (Figure \ref{fig: realfisheyedatacollection}), looking vertically down, and capture images of the environment. Our ultimate goal is to mount these cameras at challenging intersections for navigation of autonomous vehicles, and use the proposed method to register them in a prior map. We use an iPhone to provide us an approximate GPS location (without the orientation) of the fisheye camera, which is used to limit the search space in the prior map. We use a high accuracy RTK-GPS (uBlox ZED F9P GNSS + uBlox antenna ANN-MB-00) unit to measure GPS co-ordinates  of distinctive corners on road markings that can be used for quantifying the accuracy of camera localization (Figure \ref{fig: reprojectionErrors}).
\vspace{-10pt}

\subsubsection{Results} 
\begin{figure*}[!ht]
  \centering
  \subfloat[Location 1 - Projection of LiDAR points (cyan) using PnP Estimate]{\includegraphics[width=0.45\textwidth]{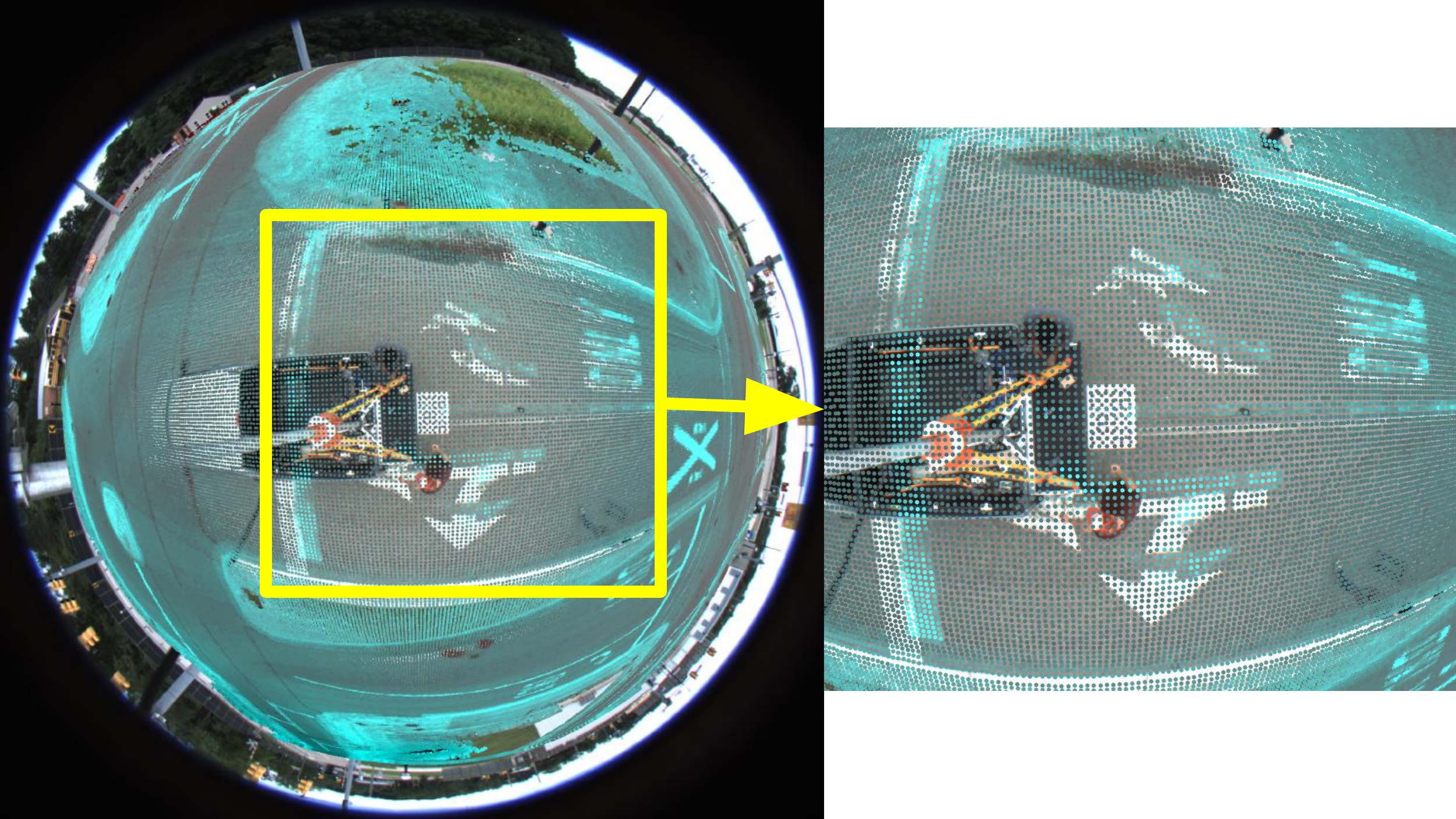}\label{fig: misalignmentinprojection_rdloc1}}
  \quad
  \subfloat[Location 1 - Projection of LiDAR points (cyan) using MI Estimate]{\includegraphics[width=0.45\textwidth]{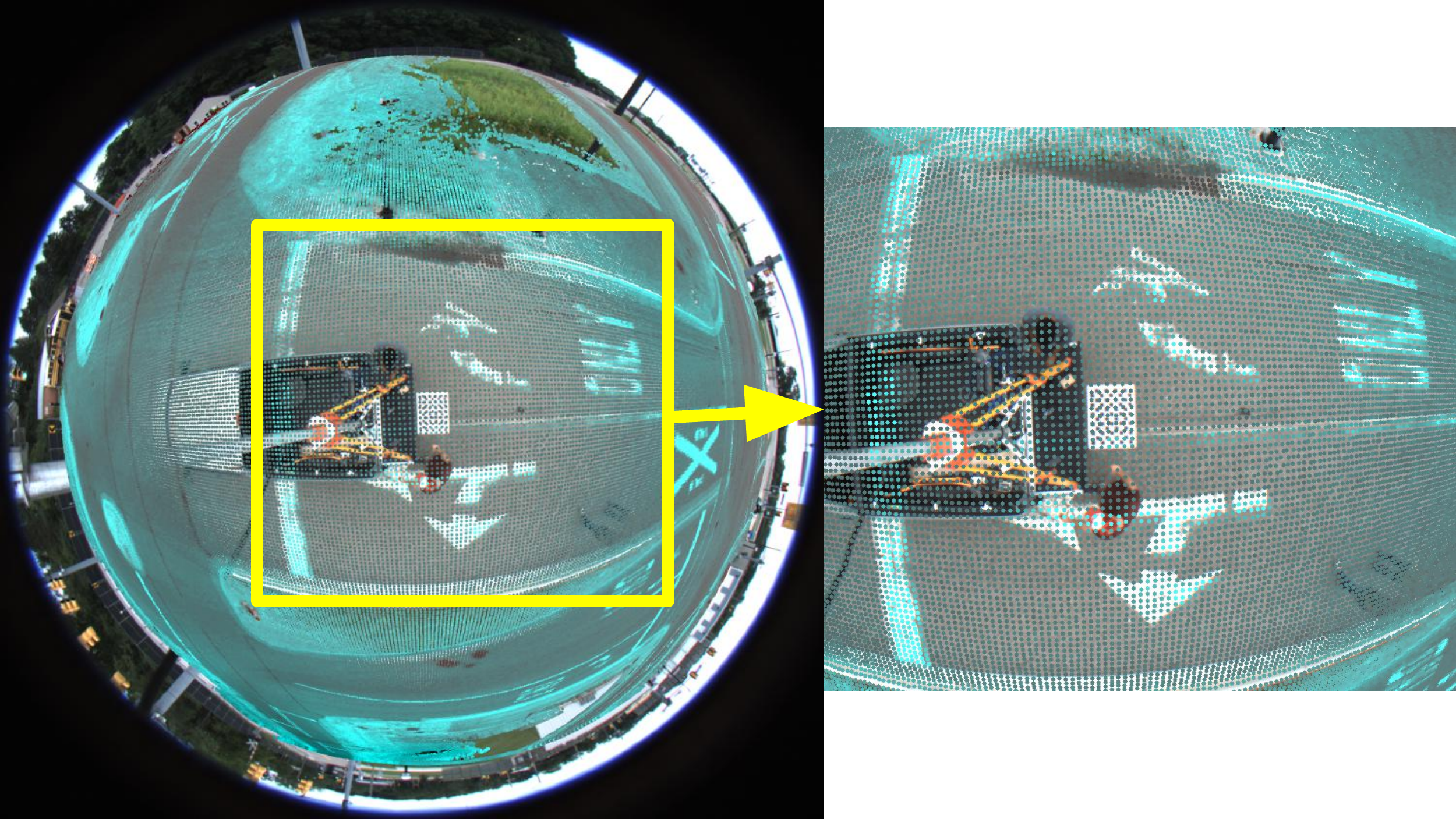}\label{fig: properalignment_rdloc1}}
  
  \subfloat[Location 2 - Projection of LiDAR points (cyan) using PnP Estimate]{\includegraphics[width=0.45\textwidth]{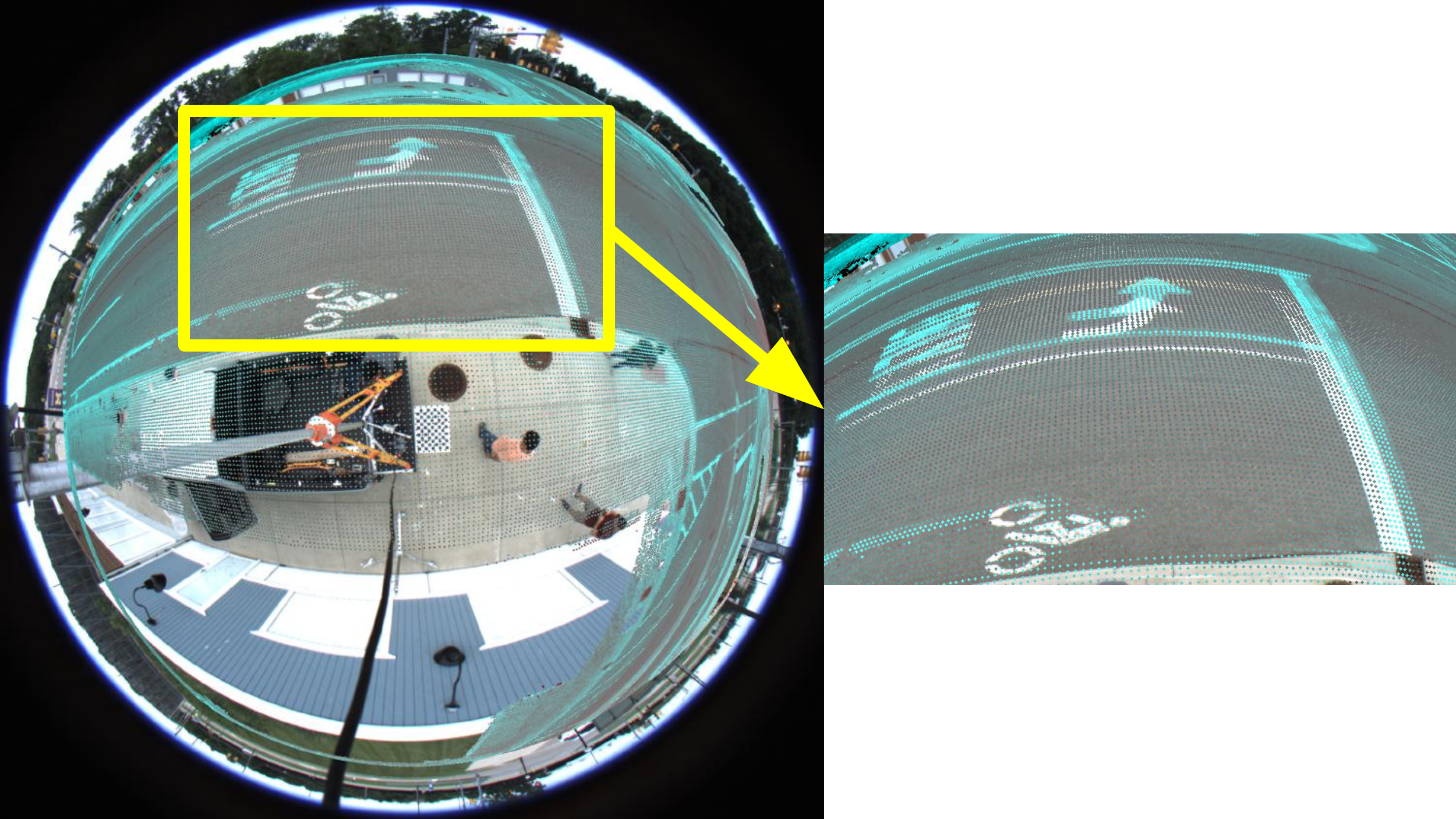}\label{fig: misalignmentinprojection_rdloc2}}
  \quad
  \subfloat[Location 2 - Projection of LiDAR points (cyan) using MI Estimate]{\includegraphics[width=0.45\textwidth]{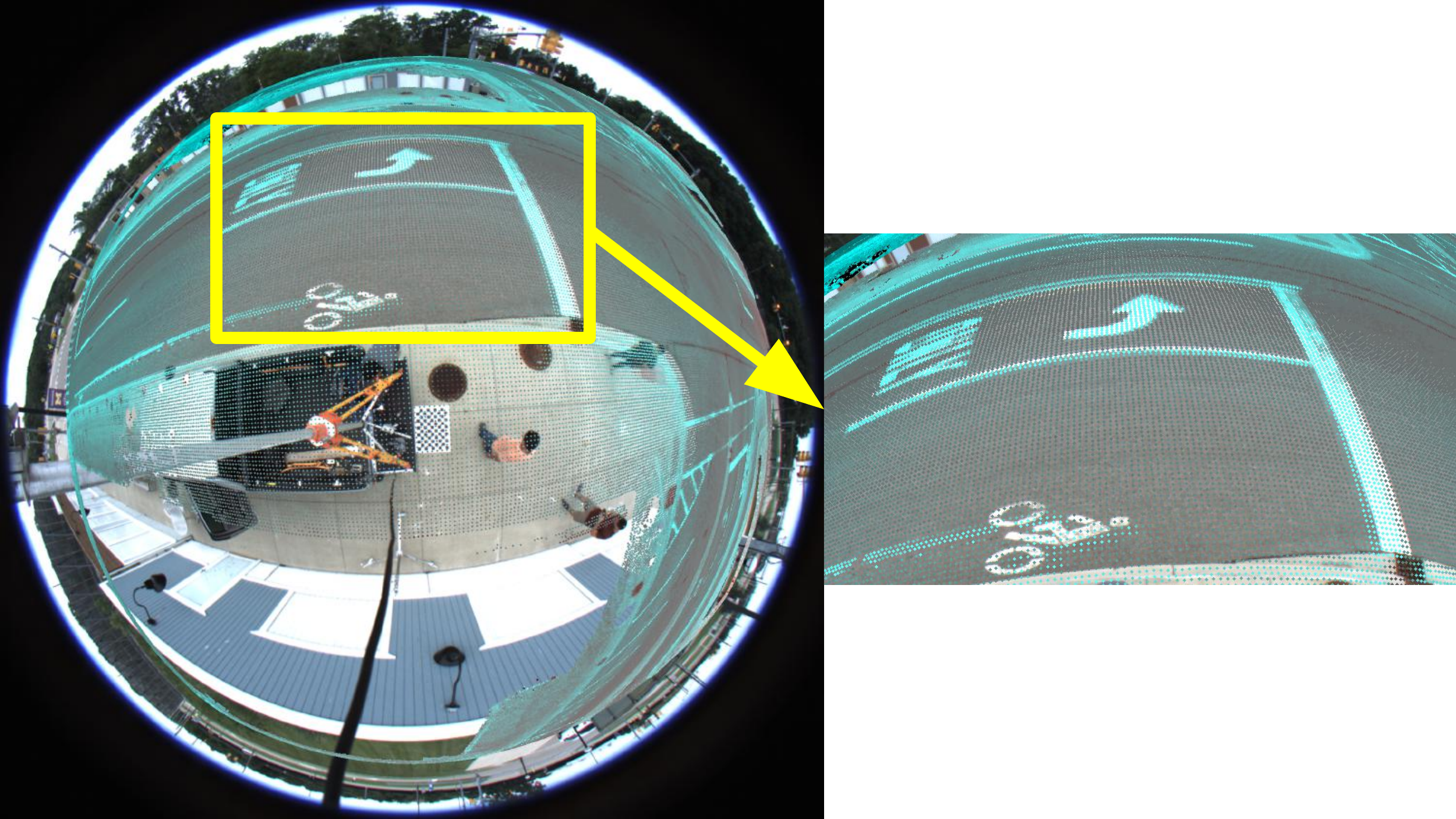}\label{fig: properalignment_rdloc2}}
  
  \caption{\textbf{Real World Experiments: }Projection of 3D-LiDAR ground points (cyan) on to fisheye camera using the initial camera pose (PnP Estimate) (Figure \ref{fig: misalignmentinprojection_rdloc1}, \ref{fig: misalignmentinprojection_rdloc2}) and MI based refinement of initial camera pose (Figure \ref{fig: properalignment_rdloc1}, \ref{fig: properalignment_rdloc2}). The misalignment of LiDAR points visible in highlighted areas in Figures \ref{fig: misalignmentinprojection_rdloc1}, \ref{fig: misalignmentinprojection_rdloc2}, are minimized in Figures \ref{fig: properalignment_rdloc1}, \ref{fig: properalignment_rdloc2}} 
  \label{fig: misalignmentvsalignmentsimulateddata_rdloc}
  \vspace{-15pt}
\end{figure*}

\begin{figure*}[!ht]
  \centering
  \subfloat[Location1 - Average Reprojection Error with PnP = \textcolor{red}{20.52} pixel, and with Maximization of MI = \textcolor{blue}{9.12} pixel]{\includegraphics[width=0.45\textwidth]{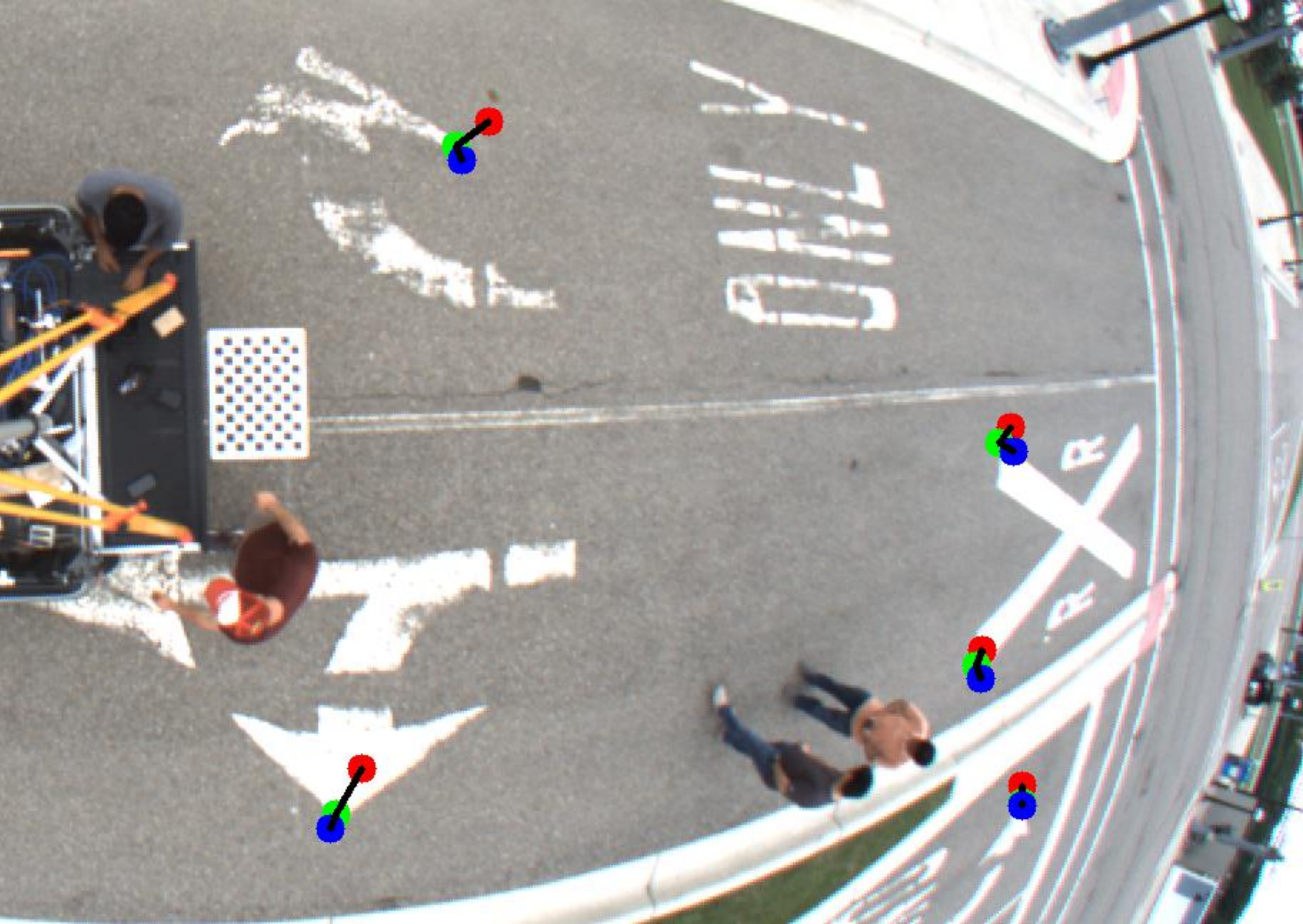}\label{fig: repErrLoc1}}
  \quad
  \subfloat[Location2 - Average Reprojection Error with PnP = \textcolor{red}{26.40} pixel, and with Maximization of MI = \textcolor{blue}{13.11} pixel]{\includegraphics[width=0.45\textwidth]{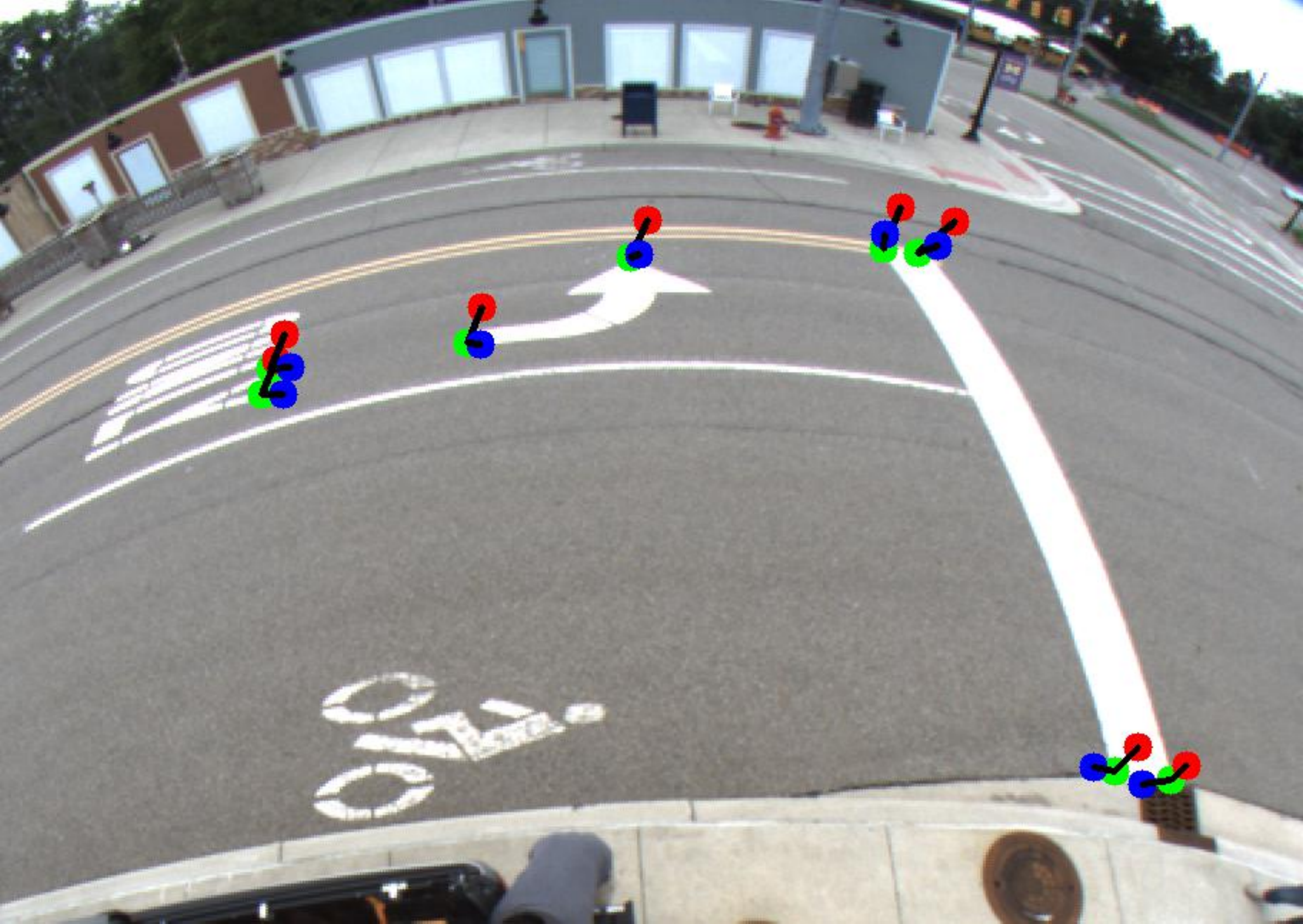}\label{fig: repErrLoc2}}
  \caption{\textcolor{green}{Green Circle} - Hand annotated corner point whose GPS location was measured using a high accuracy RTK-GPS unit, \textcolor{red}{Red Circle} - Projection of corner point's position onto Fisheye Image using PnP estimate (Section \ref{sec: initsparsefeaturematching}), \textcolor{blue}{Blue Circle} - Projection of corner point's position onto Fisheye Image using MI estimate (Section \ref{sec: refinement_of_localization}).} 
  \label{fig: reprojectionErrors}
  \vspace{-20pt}
\end{figure*}

We present results with real world data from two different locations in Figure \ref{fig: misalignmentvsalignmentsimulateddata_rdloc} which qualitatively demonstrate the incremental improvement in camera localization using our two step approach. While the projection of LiDAR points onto the fisheye image using the initial camera localization appears misaligned in Figures \ref{fig: misalignmentinprojection_rdloc1} and \ref{fig: misalignmentinprojection_rdloc2}, the misalignment is reduced when the LiDAR points are projected using refined camera localization in Figures \ref{fig: properalignment_rdloc1} and \ref{fig: properalignment_rdloc2}. We quantify the veracity of camera localization by measuring the average reprojection error for points on the fisheye image whose GPS locations we have measured using high accuracy RTK-GPS. We manually mark these points on the fisheye image, and measure the difference between them and the reprojection of the corresponding 3D point in the prior map onto the fisheye image, using the estimated camera localization. The results presented in Figure \ref{fig: reprojectionErrors} show that the reprojection error on the fisheye image plane reduces when we refine the initial camera localization by maximizing the mutual information. 

\section{Discussion $\&$ Conclusion}
\label{sec: conclusion}
We present an approach to localize a smart infrastructure node equipped with a fisheye camera. The downward facing fisheye image is registered to a prior map, comprising of a co-registered satellite image and a ground reflectivity/height map from LiDAR-SLAM. Our two-step approach uses feature matching between the rectified fisheye image and the satellite imagery to get an initial camera pose, followed by maximization of MI between the fisheye image and 3D LiDAR map to refine the initial camera localization. Since we have only a single camera image to register against (the smart infrastructure node is static), the cost surface may not always be smooth \cite{MIGP} and therefore not differentiable - leading to the failure of gradient descent methods. Hence, we use an exhaustive grid search method to find the optimal camera pose. Such a search may be time-consuming (depending on the number of 3D LiDAR map points used for calculating MI (Equation \ref{eqn: mutualinformationdefinition}), the interval of exhaustive grid search and the available compute power), and not suitable for real-time operation. This is acceptable for our application, because we need to localize the smart infrastructure node once at install and this can be an offline process. Moreover, this method can be accelerated by use of GPUs.


\bibliography{references.bib}

\begin{thebibliography}{10}
\providecommand{\url}[1]{#1}
\csname url@samestyle\endcsname
\providecommand{\newblock}{\relax}
\providecommand{\bibinfo}[2]{#2}
\providecommand{\BIBentrySTDinterwordspacing}{\spaceskip=0pt\relax}
\providecommand{\BIBentryALTinterwordstretchfactor}{4}
\providecommand{\BIBentryALTinterwordspacing}{\spaceskip=\fontdimen2\font plus
\BIBentryALTinterwordstretchfactor\fontdimen3\font minus
  \fontdimen4\font\relax}
\providecommand{\BIBforeignlanguage}[2]{{%
\expandafter\ifx\csname l@#1\endcsname\relax
\typeout{** WARNING: IEEEtran.bst: No hyphenation pattern has been}%
\typeout{** loaded for the language `#1'. Using the pattern for}%
\typeout{** the default language instead.}%
\else
\language=\csname l@#1\endcsname
\fi
#2}}
\providecommand{\BIBdecl}{\relax}
\BIBdecl

\bibitem{vora2020aerial}
A.~Vora, S.~Agarwal, G.~Pandey, and J.~McBride, ``Aerial imagery based lidar
  localization for autonomous vehicles,'' 2020.

\bibitem{maxar}
\BIBentryALTinterwordspacing
M.~Technologies. (2020). [Online]. Available: \url{https://www.maxar.com/}
\BIBentrySTDinterwordspacing

\bibitem{vora2019high}
A.~G. Vora, S.~Agarwal, J.~N. Hoellerbauer, and F.~Shaik, ``High definition 3d
  mapping,'' Jun.~6 2019, uS Patent App. 15/831,295.

\bibitem{Yu2020MonocularCL}
H.~Yu, W.~Zhen, W.~Yang, J.~Zhang, and S.~Scherer, ``Monocular camera
  localization in prior lidar maps with 2d-3d line correspondences,''
  \emph{2020 IEEE/RSJ International Conference on Intelligent Robots and
  Systems (IROS)}, pp. 4588--4594, 2020.

\bibitem{Wolcott2014VisualLW}
R.~W. Wolcott and R.~Eustice, ``Visual localization within lidar maps for
  automated urban driving,'' \emph{2014 IEEE/RSJ International Conference on
  Intelligent Robots and Systems}, pp. 176--183, 2014.

\bibitem{MIGP}
G.~Pandey, J.~McBride, S.~Savarese, and R.~Eustice, ``Automatic targetless
  extrinsic calibration of a 3d lidar and camera by maximizing mutual
  information,'' \emph{Twenty-Sixth AAAI Conference on Artificial
  Intelligence}, vol.~26, 01 2012.

\bibitem{Viswanathan}
A.~Viswanathan, B.~R. Pires, and D.~Huber, ``Vision based robot localization by
  ground to satellite matching in gps-denied situations,'' in \emph{2014
  IEEE/RSJ International Conference on Intelligent Robots and Systems}, 2014,
  pp. 192--198.

\bibitem{Lowe:2004:DIF:993451.996342}
\BIBentryALTinterwordspacing
D.~G. Lowe, ``Distinctive image features from scale-invariant keypoints,''
  \emph{Int. J. Comput. Vision}, vol.~60, no.~2, pp. 91--110, Nov. 2004.
  [Online]. Available:
  \url{http://dx.doi.org/10.1023/B:VISI.0000029664.99615.94}
\BIBentrySTDinterwordspacing

\bibitem{MeiRives}
C.~Mei and P.~Rives, ``Single view point omnidirectional camera calibration
  from planar grids,'' in \emph{Proceedings 2007 IEEE International Conference
  on Robotics and Automation}, 2007, pp. 3945--3950.

\bibitem{opencv_library}
G.~Bradski, ``{The OpenCV Library},'' \emph{Dr. Dobb's Journal of Software
  Tools}, 2000.

\bibitem{boli}
B.~Li, L.~Heng, K.~Koser, and M.~Pollefeys, ``A multiple-camera system
  calibration toolbox using a feature descriptor-based calibration pattern,''
  in \emph{2013 IEEE/RSJ International Conference on Intelligent Robots and
  Systems}, 2013, pp. 1301--1307.

\bibitem{sarlin20superglue}
\BIBentryALTinterwordspacing
P.-E. Sarlin, D.~DeTone, T.~Malisiewicz, and A.~Rabinovich, ``{SuperGlue}:
  Learning feature matching with graph neural networks,'' in \emph{CVPR}, 2020.
  [Online]. Available: \url{https://arxiv.org/abs/1911.11763}
\BIBentrySTDinterwordspacing

\bibitem{P3P}
R.~Haralick, D.~Lee, K.~Ottenburg, and M.~Nolle, ``Analysis and solutions of
  the three point perspective pose estimation problem,'' in \emph{Proceedings.
  1991 IEEE Computer Society Conference on Computer Vision and Pattern
  Recognition}, 1991, pp. 592--598.

\bibitem{RANSAC}
\BIBentryALTinterwordspacing
M.~A. Fischler and R.~C. Bolles, ``Random sample consensus: A paradigm for
  model fitting with applications to image analysis and automated
  cartography,'' \emph{Commun. ACM}, vol.~24, no.~6, p. 381–395, Jun. 1981.
  [Online]. Available: \url{https://doi.org/10.1145/358669.358692}
\BIBentrySTDinterwordspacing

\bibitem{paulviola}
P.~Viola and W.~Wells, ``Alignment by maximization of mutual information,''
  vol.~24, 01 1995, pp. 16--23.

\bibitem{MaesMI}
F.~Maes, A.~Collignon, D.~Vandermeulen, G.~Marchal, and P.~Suetens,
  ``Multimodality image registration by maximization of mutual information,''
  \emph{IEEE Transactions on Medical Imaging}, vol.~16, no.~2, pp. 187--198,
  1997.

\bibitem{roadrunner}
\BIBentryALTinterwordspacing
Mathworks, ``{Roadrunner},'' 2019. [Online]. Available:
  \url{https://www.mathworks.com/products/roadrunner.html}
\BIBentrySTDinterwordspacing

\bibitem{unrealeditor}
\BIBentryALTinterwordspacing
U.~Engine, ``{Unreal Editor},'' 2016. [Online]. Available:
  \url{https://docs.unrealengine.com/4.26/en-US/Basics/UI/}
\BIBentrySTDinterwordspacing

\end{thebibliography}
\bibliographystyle{IEEEtran}

\end{document}